\documentclass[10pt,twocolumn,letterpaper]{article}

\usepackage{wacv}
\usepackage{times}
\usepackage{epsfig}
\usepackage{graphicx}
\usepackage{amsmath}
\usepackage{amssymb}

\wacvfinalcopy 

\def\wacvPaperID{530} 

\ifwacvfinal\fi
\begin{document}

\title{Cross-modal Scene Graph Matching for
Relationship-aware \\Image-Text Retrieval}

\author{Sijin Wang\textsuperscript{1,2}, Ruiping Wang\textsuperscript{1,2}, Ziwei Yao\textsuperscript{1,2}, Shiguang Shan\textsuperscript{1,2}, Xilin Chen\textsuperscript{1,2} \\
$^1$Key Laboratory of Intelligent Information Processing of Chinese Academy of Sciences (CAS), \\Institute of Computing Technology, CAS, Beijing, 100190, China\\
$^2$University of Chinese Academy of Sciences, Beijing, 100049, China \\
{\tt\small \{sijin.wang, ziwei.yao\}@vipl.ict.ac.cn,} {\tt\small \{wangruiping, sgshan, xlchen\}@ict.ac.cn}
}

\maketitle
\ifwacvfinal\thispagestyle{empty}\fi

\begin{abstract}
Image-text retrieval of natural scenes has been a popular research topic. Since image and text are heterogeneous cross-modal data, one of the key challenges is how to learn comprehensive yet unified representations to express the multi-modal data. A natural scene image mainly involves two kinds of visual concepts, objects and their relationships, which are equally essential to image-text retrieval. Therefore, a good representation should account for both of them. In the light of recent success of scene graph in many CV and NLP tasks for describing complex natural scenes, we propose to represent image and text with two kinds of scene graphs: visual scene graph (\textbf{VSG}) and textual scene graph (\textbf{TSG}), each of which is exploited to jointly characterize objects and relationships in the corresponding modality. The image-text retrieval task is then naturally formulated as cross-modal scene graph matching. Specifically, we design two particular scene graph encoders in our model for VSG and TSG, which can refine the representation of each node on the graph by aggregating neighborhood information. As a result, both object-level and relationship-level cross-modal features can be obtained, which favorably enables us to evaluate the similarity of image and text in the two levels in a more plausible way. We achieve state-of-the-art results on Flickr30k and MSCOCO, which verifies the advantages of our graph matching based approach for image-text retrieval.
\end{abstract}


\begin{figure*}

\setlength{\abovecaptionskip}{-0.5cm}
\setlength{\belowcaptionskip}{-0.5cm}
\centering\includegraphics[width=18cm, height=7cm]{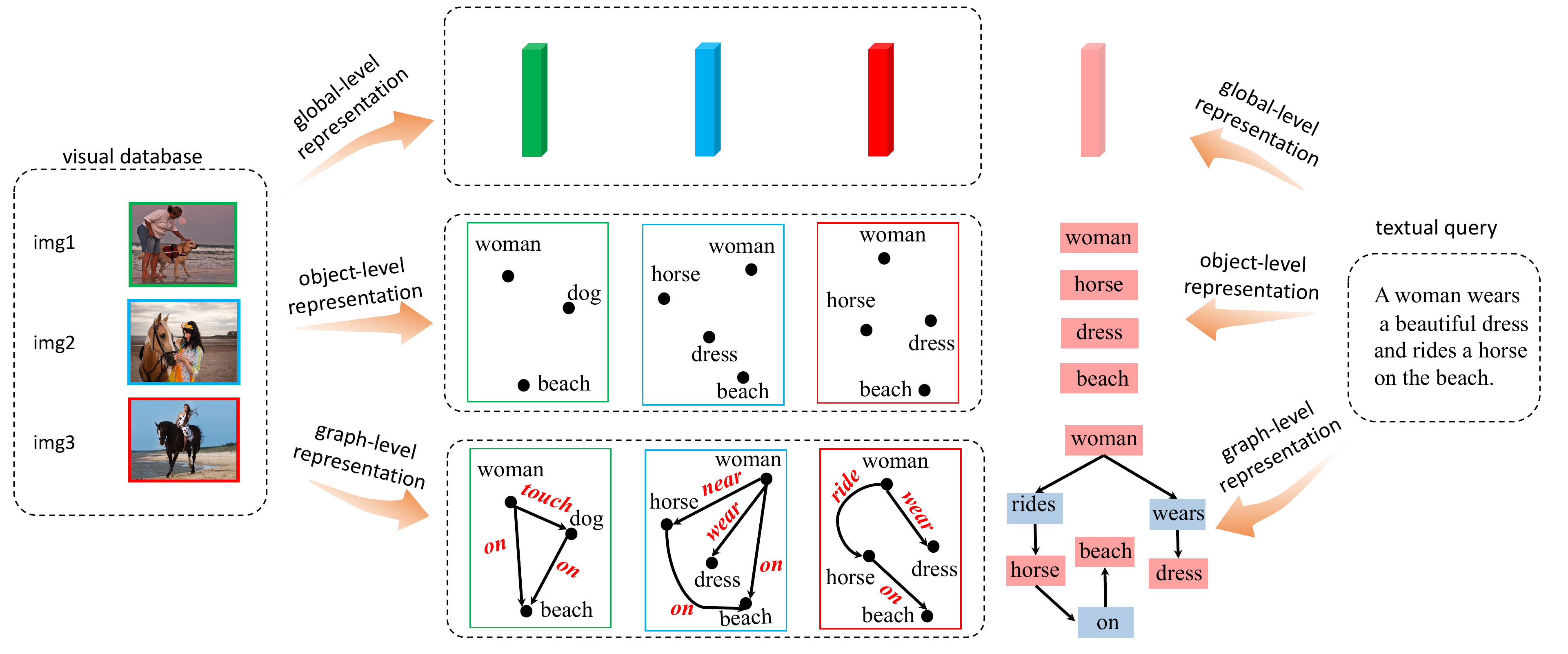}
\caption{Three different image-text retrieval frameworks. The framework on the top 
uses global representations to present images and text for matching. The middle one extracts objects in the image and text for detailed matching. The bottom one (ours) captures both objects and their relationships from the image and text with two graphs for two levels matching.}
\label{fig:f1}
\end{figure*}
\vspace{-0.3cm}
\section{Introduction}
\label{sec:intro}

Visual media 
and natural language are the two most prevalent information coming in different modalities in our daily life. To achieve artificial intelligence on computers, it is essential to enable computers to understand, match, and transform such cross-modal data. Image-text cross-modal retrieval is thus one of the challenging research topics, where given a query of one modality (an image or a text sentence), it aims to retrieve the most similar samples from the database in another modality.   
The key challenge here is how to match the cross-modal data by understanding their contents and measuring their semantic similarity, especially when there are multiple objects in the cross-modal data. 


To address this task, many approaches have been proposed. As shown in the top of Fig.\ref{fig:f1}, early approaches \cite{kiros2014unifying,faghri2018vse++,wang2017adversarial,wang2016learning,zhang2018deep} use global representations to express the whole image and sentence, which ignore the local details. Such approaches work well on simple cross-modal retrieval scenario that contains only a single object, but are not satisfactory for more realistic cases that involve complex natural scenes. Recent studies \cite{karpathy2014deep,karpathy2015deep,huang2017instance,huang2018learning,lee2018stacked} pay attention to local detailed matching by detecting objects in both images and text, and have gained certain improvements over previous works, which is described in the middle of Fig.\ref{fig:f1}.  

However, a natural scene contains not only several objects but also their relationships \cite{johnson2015image}, which are equally important to image-text retrieval. For example, three images in the left of Fig.\ref{fig:f1} contain similar objects. The \textit{``dog''} in $img1$ can distinguish this image from the other two, while $img2$ and $img3$ contain the same objects, including \textit{``woman''}, \textit{``horse''}, \textit{``beach''} and \textit{``dress''}. To discriminate such two images, the relationships play an essential role. Clearly, the \textit{``woman''} in $img2$ is \textit{``standing next to''} the horse while the \textit{``woman''} in $img3$ is \textit{''riding on''} the horse. Similarly, there are also semantic relationships between textual objects in a sentence after syntactic analysis, such as \textit{``woman-wears-dress''}, \textit{``woman-rides-on-horse''} in the text query in Fig.\ref{fig:f1}.

With more recent research topics focusing on the objects and relationships in the image scene, scene graphs \cite{johnson2015image} are proposed to model the objects and relationships formally and have quickly become a powerful tool used in high-level semantic understanding tasks \cite{li2017scene,Yao2018ExploringVR,Johnson2018ImageGF,Wang2018NeighbourhoodWR,Yang2018AutoEncodingSG}. A scene graph consists of many nodes and edges, in which each node represents an object, and each edge indicates the relationship between the two nodes it connects. To represent the image and text comprehensively in the image-text retrieval task, we organize the objects and the relationships into scene graphs for both modalities, as illustrated in the bottom of Fig.\ref{fig:f1}.
We introduce a visual scene graph (VSG) and a textual scene graph (TSG) to represent images and text, respectively, converting the conventional image-text retrieval problem to the matching of two scene graphs.

To be specific, we extract objects and relationships from the image and text to form the VSG and TSG, and design a so-called Scene Graph Matching (\textbf{SGM}) model, where two tailored graph encoders encode the VSG and TSG into the visual feature graph and the textual feature graph. The VSG encoder is a Multi-modal Graph Convolution Network (MGCN), which enhances the representations of each node on the VSG by aggregating useful information from other nodes and updates the object and relationship features in different manners. The TSG encoder contains two different bi-GRUs aiming to encode the object and relationship features, respectively. After that, both object-level and relationship-level features are learned in each graph, and the two feature graphs corresponding to 
two modalities can be finally matched at two levels in a more plausible way.

To evaluate the effectiveness of our approach, we conduct image-text retrieval experiments on two challenging datasets, Flickr30k \cite{Young2014FromID} and MSCOCO \cite{Lin2014MicrosoftCC}. The results show that the performance of our approach significantly outperforms state-of-the-art methods and validates the importance of relationships for image-text retrieval.

\begin{figure*}[t]
\setlength{\abovecaptionskip}{-0.5cm}
\setlength{\belowcaptionskip}{-0.5cm}
\centering\includegraphics[width=18cm]{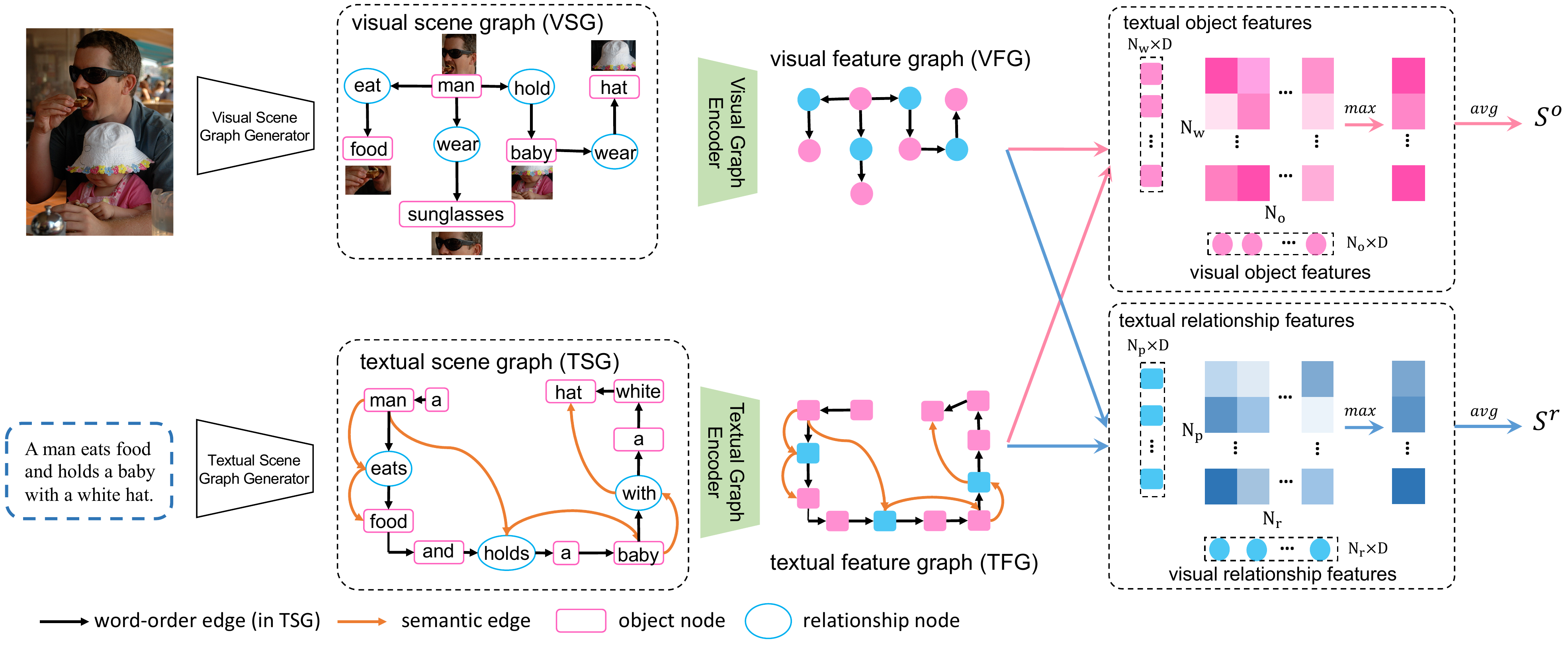}
\caption{
The architecture of our method. The image and sentence are parsed into a VSG and TSG by two graph generators. Then two encoders encode them into feature graphs, which are matched at object and relationship levels at last. Note that VSG and TSG have two types of nodes and TSG contains two kinds of edges that are explained in the legend.
}
\label{fig:f2}
\end{figure*}

\section{Related Works}
\textbf{Image-Text Retrieval.} 
Image-text retrieval task has become a popular research topic in recent years. Several excellent works \cite{kiros2014unifying,faghri2018vse++,socher2014grounded,wang2017adversarial,karpathy2014deep,karpathy2015deep,huang2017instance,huang2018learning,lee2018stacked,Klein2015AssociatingNW,zhang2018deep,gu2018look} are introduced to address this task, which can be divided into two groups: i) global representation based methods and ii) local representation based methods.

Global representation based methods \cite{faghri2018vse++,wang2017adversarial,wang2016learning,zhang2018deep,frome2013devise,kiros2014unifying} usually consist of an image encoder (\eg CNN) and a sentence encoder (\eg RNN) to extract a global feature of the image and sentence, respectively. Then, a metric is devised to measure the similarity of a couple of features in different modalities. Frome \etal \cite{frome2013devise} proposed a deep visual semantic embedding model that uses CNN to extract the visual representations from the full image and Skip-Gram \cite{mikolov2013efficient} to obtain the representation of the semantic labels. 
Similarly, Kiros \etal \cite{kiros2014unifying} use LSTM to encode the full sentence and the triplet loss to make the matched image-sentence pair closer than the unmatched pairs in the embedding space. Wehrmann \etal \cite{wehrmann2018bidirectional} designed an
efficient character-level inception module which encodes textual features by convolving raw characters in the sentence. Faghri \etal \cite{faghri2018vse++} produce significant gains in retrieval performance by introducing hard negatives mining into triplet loss. 

To be more detailed, local representation based methods \cite{karpathy2014deep,karpathy2015deep,huang2017instance,huang2018learning,lee2018stacked} that focus on the local alignment between images and sentences, have been developed recently. Karpathy \etal \cite{karpathy2014deep} extract objects from images, and match these visual objects with words in the sentences. To improve such approach, Lee \etal \cite{lee2018stacked} attend more important fragments (words or regions) with an attention network. Huang \etal \cite{huang2018learning} propose that semantic concepts, as well as the order of semantic concepts, are essential for image-text matching. To slove the issue of embedding polysemous instances, Song and Soleymani \cite{song2019polysemous} extract K embeddings of each image rather than injective embedding.
However, some of above methods lose sight of the relationships between objects in multi-modal data, which is also the key point for image-text retrieval. Though some of them \cite{karpathy2015deep,huang2017instance,huang2018learning,lee2018stacked} use RNNs to embed words with context, it still does not explicitly reveal the semantic relationships between textual objects. In our approach, both visual and textual objects with their relationships are explicitly captured by scene graphs. Thus, the cross-modal data can match in two levels, which is more plausible.

\textbf{Scene Graph.}  
Scene graph was first proposed by \cite{johnson2015image} for image retrieval, which describes objects, their attributes, and relationships in images with a graph. With recent breakthroughs in scene graph generation \cite{zellers2018neural,li2017scene,Xu2017SceneGG,wang2018scene,wang2018scene}, many high-level visual semantic tasks are developed, such as VQA \cite{Teney2017GraphStructuredRF}, image captioning \cite{Yang2018AutoEncodingSG,Yao2018ExploringVR,li2017scene}, and grounding referring expressions \cite{Wang2018NeighbourhoodWR}. Most of these methods benefit from the use of scene graphs to present images. On the other hand, several methods \cite{Anderson2016SPICESP,wang2018scene,schuster2015generating} are proposed to parse the sentence into a scene graph, which is applied to some cross-modal tasks \cite{Yang2018AutoEncodingSG}. In recent years, there are attempts to use graph structures to represent both visual and textual data, such as \cite{Teney2017GraphStructuredRF} that employs graphs to represent image and text questions for VQA. Distinctive from our method, their graphs are not the so-called scene graph, which contain no semantic relationships.


\section{Method}
\label{sec:med}

Given a query in one modality (a sentence query or an image query), the goal of the image-text cross-modal retrieval task is to find the most similar sample from the database in another modality. Therefore, our Scene Graph Matching (SGM) model aims to evaluate the similarity of the image-text pairs by dissecting the input image and text sentence into scene graphs. The framework of SGM is illustrated in Fig.\ref{fig:f2}, which consists of two branches of networks. In the visual branch, the input image is represented into a visual scene graph (VSG) and then encoded into the visual feature graph (VFG). Simultaneously, the sentence is parsed into a textual scene graph (TSG) and then encoded into the textual feature graph (TFG) in the textual branch. Finally, the model collects object features and relationship features from the VFG and TFG and calculates the similarity score at the object-level and relationship-level, respectively. The architectures of the submodules of SGM will be detailed in the following subsections.

\begin{figure}
\setlength{\abovecaptionskip}{-0.5cm}
\setlength{\belowcaptionskip}{-0.5cm}
\centering\includegraphics[width=8.2cm]{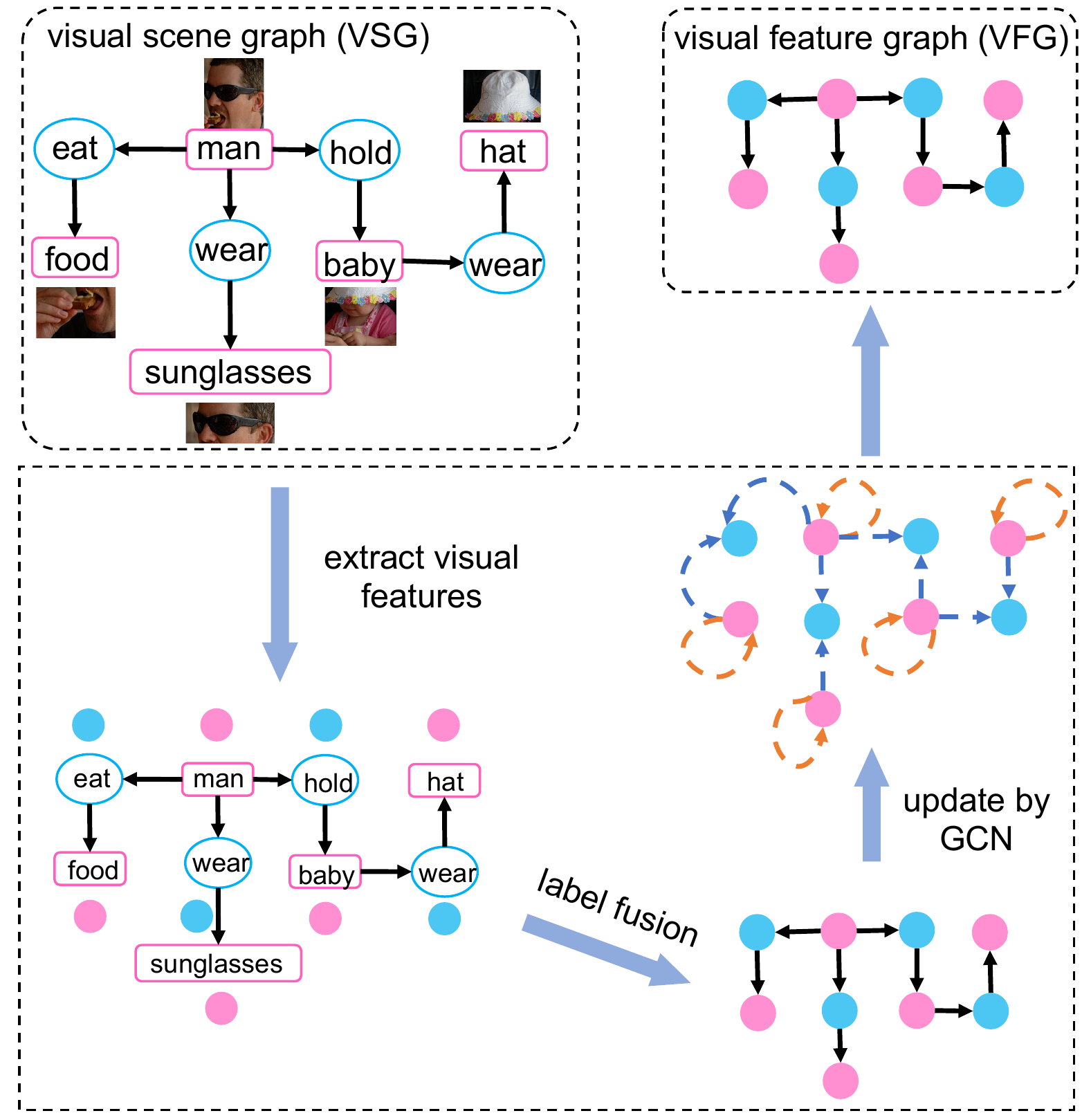}
\caption{
The framework of the VSG encoder. The corresponding image region of each node is embedded into a feature vector by the visual feature extractor. Then the visual feature and the word label of each node are fused by the multi-modal fusion layer. Finally, the graph is encoded by a GCN, where each node collects information and updates its representation as indicated by the dashed arrow, and yield the visual feature graph as output.
}
\label{fig:vsg-encoder}
\end{figure}

\subsection{Visual Feature Embedding}
\subsubsection{Visual Scene Graph Generation}
Given a raw image, the visual scene graph is generated by an off-the-shelf scene graph generation method, such as MSDN \cite{li2017scene} and NeuralMotifs \cite{zellers2018neural}.
We represent a visual scene graph as $G=\{V, E\}$, where $V$ is the node-set, and $E$ is the edge-set. There are two types of nodes in our visual scene graph, as shown in Fig.\ref{fig:f2}. The pink rectangles denote object nodes, each of which corresponds to a region of the image. The ellipses in light blue are relationship nodes, each of which connects two object nodes by directed edges. Additionally, each node has a category label, such as \textit{``man'',``hold''}.  

Concretely, suppose there are $N_o$ object nodes and $N_r$ relationship nodes in a VSG. The object nodes set can be represented as $O=\{o_i | i=1, 2,\dots,N_o\}$. 
The set of relationship nodes is $R=\{r_{ij}\}\subseteq{O}\times{O}$, where $|R|=N_r$, and $r_{ij}$ is the relationship of $o_i$ and $o_j$. The label of $o_i$ and $r_{ij}$ can be represented by one-hot vectors, $\textbf{l}_{o_i}$ and $\textbf{l}_{r_{ij}}$. 

\vspace{-0.5cm}
\subsubsection{Visual Scene Graph Encoder} 
After the generation of visual scene graph, we design a \textit{Multi-modal Graph Convolution Network} (MGCN) to learn good representations on VSGs, which includes a pre-trained visual feature extractor, a label embedding layer, a multi-modal fusion layer, and a graph convolution network, shown in Fig.\ref{fig:vsg-encoder}.

\textbf{Visual Feature Extractor.}  
The pre-trained visual feature extractor is used for encoding image regions into feature vectors, which can be pre-trained CNN networks 
or object detectors (\eg Faster-RCNN \cite{ren2015faster}). Each node in the VSG will be encoded into a $d_1$-dimension visual feature vector by the extrator. For object node $o_i$, its visual feature vector $\textbf{v}_{o_i}$ is extracted from its corresponding image region. For relationship node $r_{ij}$, its visual feature vector $\textbf{v}_{r_{ij}}$ is extracted from the union image region of $o_i$ and $o_j$.

\textbf{Label Embedding Layer.}
Each node has a word label predicted by the visual scene graph generator, which can provide the auxiliary semantic information. The label embedding layer is built to embed the word label of each node into a feature vector.
Given the one-hot vectors $\textbf{l}_{o_i}$ and $\textbf{l}_{r_{ij}}$, the embedded label features $\textbf{e}_{o_i}$ and $\textbf{e}_{r_{ij}}$ are computed as $\textbf{e}_{o_i}=\textbf{W}_o\textbf{l}_{o_i}$ and $\textbf{e}_{r_{ij}}=\textbf{W}_{r}\textbf{l}_{r_{ij}}$, where $\textbf{W}_o\in{R^{{d_2}\times{C_o}}}$ and $\textbf{W}_r\in{R^{{d_2}\times{C_r}}}$ are trainable parameters and initialized by word2vec (we use $d_2$=300). $C_o$ is the category number of objects and $C_r$ is the category number of relationships.

\textbf{Multi-modal Fusion Layer.}
After obtaining the visual feature and label feature of each node, it is necessary to fuse them into a unified representation. Thus, a multi-modal fused feature graph is generated. Specifically, the visual feature and label feature are concatenated, then fused as
\begin{equation}
\textbf{u}_{o_i}=\tanh(\textbf{W}_u[\textbf{v}_{o_i},\; \textbf{e}_{o_i}]),
\end{equation}
\begin{equation}
\textbf{u}_{r_{ij}}=\tanh(\textbf{W}_u[\textbf{v}_{r_{ij}},\; \textbf{e}_{r_{ij}}]),
\end{equation}
where $\textbf{W}_u\in{R^{d_1\times{(d_1+d_2)}}}$ is the trainable parameter of fusion layer.

\textbf{Graph Convolution Network.} GCNs \cite{Wu2019ACS} are convolutional neural networks that can operate on graphs of any structure, which is more flexible than CNNs that can only work on grid structured
data. To encode the multi-modal fused feature graph, we adopt an $m$-layer GCN 
and propose a novel update mechanism to update two kinds of nodes in different manners. The object nodes will generate object-level features, which can be seen as the first-order features of the image. It may ruin the representation of the object node by the information from another object node or relationship node so that each object node is updated without other information from the neighborhoods. On the contrary, the relationship-level features are the second-order features of the image, so the representations of relationship nodes can be enhanced by its adjacent object nodes. Therefore, relationship nodes update by aggregating information from their neighborhoods and object nodes update from themselves, as shown by the blue and yellow dashed arrows in Fig.\ref{fig:vsg-encoder}. Concretely, given the multi-modal fused feature graph $\mathcal{G}=\{\mathcal{V},\mathcal{E}\}$ (distinguished from the raw visual scene graph $G=\{V,E\}$), the $k$-th layer of GCN is computed as
\begin{equation}
\textbf{h}_{o_i}^k=g_o(\textbf{h}_{o_i}^{k-1}), \qquad \textbf{h}_{r_{ij}}^k=g_r(\textbf{h}_{o_i}^{k-1},\textbf{h}_{r_{ij}}^{k-1},\textbf{h}_{o_j}^{k-1}),
\end{equation}
where $g_r$ and $g_o$ are fully-connected layers, followed by a tanh function. The initial hidden states are the fused features as $\textbf{h}_{o_{i}}^0=\textbf{u}_{o_i}$ and $\textbf{h}_{r_{ij}}^0=\textbf{u}_{r_{ij}}$.

Finally, the output of an $m$-layer GCN 
is an encoded visual feature graph with two kinds of vertices: $\textbf{h}_{o_i}$, $\textbf{h}_{r_{ij}}$.


\begin{figure*}[t]
\setlength{\abovecaptionskip}{-0.5cm}
\setlength{\belowcaptionskip}{-0.5cm}
\centering\includegraphics[width=18cm]{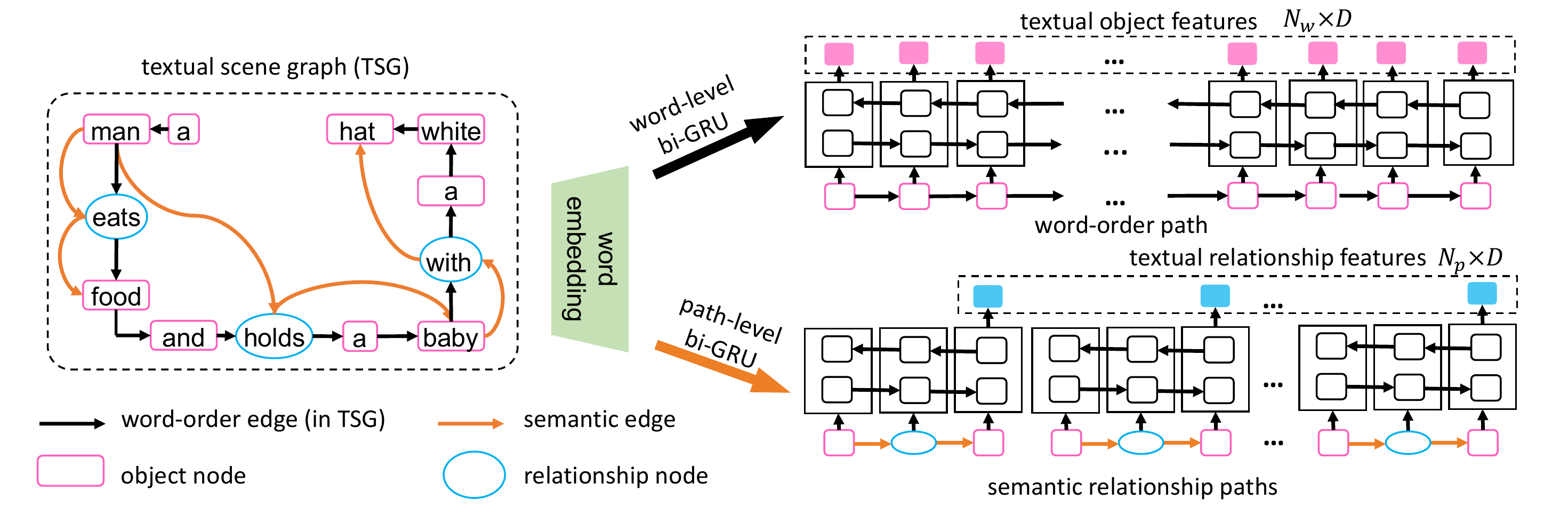}
\caption{The architecture of the textual scene graph encoder. After embedding each word into a vector by the word embedding layer, paths connected by different edges are encoded separately by word-level bi-GRU and path-level bi-GRU. }
\label{fig:f3}
\end{figure*}

\subsection{Textual Feature Embedding}
\subsubsection{Textual Scene Graph Generation}
Similar to images, a natural language sentence also describes many objects and their relationships. Therefore, the graph structure is also appropriate for representing a sentence. We organize the words of the input sentence into a textual scene graph (TSG), which includes two kinds of edges shown in Fig.\ref{fig:f3}. The black arrows indicate word-order edges, which connect words by the word order in the sentence. The brown arrows are semantic relationship edges, which are built from semantic triplets parsed by SPICE \cite{Anderson2016SPICESP}, such as \textit{``man-hold-baby''}. Due to different kinds of edges, different types of paths are formed in the graph. The path connected by word-order edges is named as the word-order path. Paths connected by semantic relationship edges are called semantic relationship paths. 

\subsubsection{Textual Scene Graph Encoder}
Similar to the processing on the VSG, a textual scene graph encoder is devised to extract object and relationship features from the TSG, which consists of a word embedding layer, a word-level bi-GRU encoder, and a path-level bi-GRU encoder illustrated in Fig.\ref{fig:f3}. The word-level bi-GRU encoder will encode each node along the word-order path, after which the object-level feature with context is generated at each hidden state. Due to that the semantic relationship edges break the limitation of the grammatical structure of the sentence, explicit relationship-level features are obtained after the path-level bi-GRU encodes along the semantic relationship paths.

Suppose there are $N_w$ words and $N_p$ semantic triplets in a sentences, its TSG will contain $N_w$ nodes, one word-order path and $N_p$ semantic relationship paths. Firstly, each word $w_i$ is embedded into a vector by the word embedding layer as $e_{w_i}=W_el_{w_i}$, where $l_{w_i}$ is the one-hot vector of $w_i$ and $W_e$ is the parameter matrix of embedding layer. We initialize $W_e$ using the same word2vec in the VSG encoder, and then learn $W_e$ during training end-to-end. Next, two kinds of paths are encoded separately by different bi-GRUs. For the word-order path, the word-level bi-GRU operates from the start word to the end as
\begin{equation}
\begin{aligned}
\setlength\abovedisplayskip{9.5pt plus 3pt minus 7pt}
&\stackrel{\longrightarrow}{\textbf{h}_{w_i}}=\stackrel{\longrightarrow}{GRU_w}(\textbf{e}_{w_i},\; \stackrel{\longrightarrow}{\textbf{h}_{w_{i-1}}}),\\ 
&\stackrel{\longleftarrow}{\textbf{h}_{w_i}}= \stackrel{\longleftarrow}{GRU_w}(\textbf{e}_{w_i}, \; \stackrel{\longleftarrow}{\textbf{h}_{w_{i+1}}}), \qquad 
i\in[1, \; N_w],
\end{aligned}
\end{equation}  
where $\stackrel{\longrightarrow}{\textbf{h}_{w_i}}$ and $\stackrel{\longleftarrow}{\textbf{h}_{w_i}}$ are the hidden vectors of $w_i$ from two directions. Finally, the word node feature is gained as $\textbf{h}_{w_i}=(\stackrel{\longrightarrow}{\textbf{h}_{w_i}}+\stackrel{\longleftarrow}{\textbf{h}_{w_i}})/2$, which is regarded as a textual object feature.
For the $N_p$ semantic relationship paths, each of them is encoded by the path-level bi-GRU as 
\begin{equation}
\textbf{h}_{p_i}=\frac{\stackrel{\longrightarrow}{GRU_p}(path_i)+\stackrel{\longleftarrow}{GRU_p}(path_i)}{2},  i\in[1, \; N_p]
\end{equation}
$\textbf{h}_{p_i}$ is the last hidden state feature of $i$-th semantic relationship path, which is also a relationship feature of the TSG.

\subsection{Similarity Function}
To measure the similarity of two encoded graphs in different modalities, we need a similarity function. Since there are two levels of features in each graph, we match them respectively. Take object features for example, let's suppose there are $N_o$ and $N_w$ object features in the visual and textual feature graphs, each of which is a $D$-dimension vector. Inspired by \cite{karpathy2015deep}, we define the similarity score of two feature vectors $h_i$ and $h_j$ as $h_i^Th_j$. We calculate the similarity scores of all visual and textual object nodes, and then get a $N_w \times N_o$ score matrix, as shown in Fig.\ref{fig:f2}. We find the maximum value of each row, which means for every textual object, the most related visual object among $N_o$ visual objects is picked up. At last, we average them as the object-level score of two graphs. The relationship-level score is calculated in the same way.
The above process can be formulated as
\vspace{-0.1cm}
\begin{equation}
\setlength\abovedisplayskip{7pt plus 3pt minus 7pt}
S^{o}=({\sum\nolimits_{t=1}^{N_w}\max\limits_{i\in[1,\; N_o]}\textbf{h}_{w_t}^T\textbf{h}_{o_i}})/{N_w},
\end{equation}
\begin{equation}
S^{r}=({\sum\nolimits_{t=1}^{N_p}\max\limits_{r_{ij}\in R}\textbf{h}_{p_t}^T\textbf{h}_{r_{ij}}})/{N_p}.
\end{equation}
Finally, given a visual and textual feature graph, the similarity score is defined as $S=S^{o}+S^{r}$.

\subsection{Loss Function}
\label{sec:lf}
Triplet loss is commonly used in the image-text retrieval task, which constrains the similarity score of the matched image-text pairs larger than the similarity score of the unmatched ones by a margin, formulated as
\begin{equation}
\begin{aligned}
L(k,l)&=\sum\limits_{\hat{l}}\max(0,m-S_{kl}+S_{k\hat{l}}) \\
&+\sum\limits_{\hat{k}}\max(0,m-S_{kl}+S_{\hat{k}l}).
\end{aligned}
\end{equation}

$m$ is a margin parameter, image $k$ and sentence $l$ are corresponding pairs in a mini-batch, image $k$ and sentence $\hat{l}$ are non-corresponding pairs, so are image $\hat{k}$ and sentence $l$. Faghri \etal \cite{faghri2018vse++} discovered that using the hardest negative in a mini-batch during training rather than all negatives samples can boost performance. Therefore, we follow \cite{faghri2018vse++} in this study and define the loss function as 
\begin{equation}
\begin{aligned}
\label{eq_l+}
L_+(k,l)&=\max(0,m-S_{kl}+S_{kl^{'}})\\
& + \max(0,m-S_{kl}+S_{k^{'}l}),
\end{aligned}
\end{equation}
where $l^{'}=\arg\max_{j\neq{l}}S_{kj}$ and $k^{'}=\arg\max_{j\neq{k}}S_{jl}$ are the hardest negatives in the mini-batch.

\section{Experiments}
In the subsections, we will clarify the datasets and evaluation metrics we use for experiments. Then we give the details on implementation and show the 
experiment results. 
\subsection{Datasets and Evaluation Metrics}
Flickr30k \cite{Young2014FromID} and MSCOCO \cite{Lin2014MicrosoftCC} are two commonly used datasets in the image-text retrieval task, which contain $31,783$ and $123,287$ images respectively. Both of them have five text captions for each image. Following \cite{faghri2018vse++,lee2018stacked}, we split Flickr30k as $1,000$ images for validation, $1,000$ images for testing and the rest for training. For MSCOCO, we split $5,000$ images for validation, $5,000$ images for testing and $113,287$ images for training.

To demonstrate the effectiveness of our approach, we conduct caption retrieval and image retrieval experiments on Flickr30k and MSCOCO datasets. We adopt two universal metrics, R@k and Medr. R@k is the percentage of queries whose ground-truth is ranked within top K. Medr is the median rank of the first retrieved ground-truth.

\subsection{Implementation Details}
\label{sec:im_de}
\textbf{Visual Scene Graph Generation.} We use NeuralMotifs \cite{zellers2018neural} as visual scene graph generators, which can recognize 150 categories of objects and 50 categories of relationships. We pick the top $N_o$ ($N_o$=36) objects with bounding boxes and $N_r$ ($N_r$=25) relationships between them sorted by classification confidence. 

\textbf{Pre-trained Visual Feature Extractor.} After parsing the input image into a scene graph, some objects are detected with bounding boxes. We need to transform these image regions into real-valued features. We can use the features extracted by the scene graph generator. However, most of the recent scene graph generators limit to recognize 150 categories of objects and 50 categories of relationships. Since text descriptions are rich and open-vocabulary, such visual features are not expressive enough. Moreover, most of the recent scene graph generators, including Neural Motifs \cite{zellers2018neural}, use VGG \cite{Simonyan15} as the backbone. In order to make a fair comparison with some state-of-the-art approaches that use Resnet \cite{he2016deep}, we prefer a feature extractor with ResNet backbone. Therefore, we take the Faster-RCNN \cite{ren2015faster} detector, which is trained on Visual Genome dataset \cite{krishnavisualgenome} by 1600 object classes in \cite{Anderson2018BottomUpAT}. 
We use the $2048$-dimension feature vector after RoI pooling.

\textbf{Parameters Setting.} Our SGM is implemented with Pytorch platform\footnote{The source codes will be released to the public soon.}. The output dimension of visual and textual scene graph encoder is 1024. The number of layers of GCN in visual scene graph encoder is 1. The margin $m$ in loss function is set to 0.2. We use Adam \cite{kingma2014adam} optimizer with a mini-batch size of 200 to train our model. The initial learning rate is 0.0005 for MSCOCO and 0.0002 for Flickr30k. 

\begin{table}

   \begin{footnotesize}
      \begin{center}
         \caption{Evaluation of different variants of our model on Flickr30k}
         \label{table:ablation_flickr}
         \begin{tabular}{l|c|c|c|c|c|c}
            \hline
            \multicolumn{1}{l|}{model} &\multicolumn{3}{c|}{caption retrieval} & \multicolumn{3}{c}{image retireval} \\ \hline
             {}              & R@1  & R@5  & R@10  & R@1& R@5 & R@10  \\ \hline
            OOM \textbf{w/o} TCxt& 52.7 & 81.8 & 90.3 & 44.6& 72.2& 80.9  \\ \hline
            OOM  & 67.6 & 89.7 & 94.5 & 48.6& 75.6& 83.8   \\ \hline
            OOM \textbf{w} VRel   & 65.5 & 89.5 & 95.4  & 50.0& 77.4& 85.0    \\ \hline
            OOM \textbf{w} TRel     & \textbf{71.8} & 91.4 & \textbf{96.1}  & 51.0& 79.5& \textbf{86.8}  \\ \hline
            SGM  & \textbf{71.8} & \textbf{91.7} & 95.5   & \textbf{53.5}& \textbf{79.6}& 86.5  \\ \hline
         \end{tabular}  
      \end{center}
   \end{footnotesize}
   \vspace{-0.3cm}
\end{table}


\begin{table*}[t]
\begin{center}
\caption {Comparisons of state-of-the-art models on Flickr30k in cross-modal retrieval.}
\label{sofa_flickr}
\begin{tabular}{c|c|c|c|c|c|c|c|c}
\hline
\multicolumn{1}{c|}{model} & \multicolumn{4}{c|}{caption retrieval} & \multicolumn{4}{c}{image retrieval} \\ \hline
\multicolumn{1}{c|}{}      & R@1      & R@5     & R@10    & Medr    & R@1     & R@5     & R@10    & Medr   \\ \hline
VSE++ \cite{faghri2018vse++} & 52.9     & 80.5    & 87.2    & 1.0     & 39.6    & 70.1    & 79.5    & 2.0    \\ \hline
GXN \cite{gu2018look} & 56.8     & -       & 89.6    & 1.0     & 41.5    & -       & 80.1    & 2.0    \\ \hline
SCO \cite{huang2018learning}   & 55.5     & 82.0    & 89.3    & -       & 41.1    & 70.5    & 80.1    & -      \\ \hline
SCAN(t2i) AVG loss \cite{lee2018stacked} & 61.8     & 87.5    & 93.7    & -       & 45.8    & 74.4    & 83.0    & -      \\ \hline
SCAN(i2t) AVG loss \cite{lee2018stacked}  & 67.9    & 89.0   & 94.4    & -     & 43.9    & 74.2    & 82.8    & -    \\ \hline
Ours (SGM)                   & \textbf{71.8}     & \textbf{91.7}    & \textbf{95.5}    & \textbf{1.0}     & \textbf{53.5}    & \textbf{79.6}    & \textbf{86.5}    & \textbf{1.0}    \\ \hline
\end{tabular}
\end{center}
\vspace{-0.5cm}
\end{table*}

\begin{table*}[h]
\caption{Comparisons of state-of-the-art models on MSCOCO. 5k test images are the whole test dataset. 1k test images mean the test dataset is divided into five 1k subsets, and the results are the average performance on them. Results marked by '*' are our implementation with the published code and data.}
\label{sofa_coco}
\begin{center}
\begin{tabular}{c|c|c|c|c|c|c|c|c}
\hline
model               & \multicolumn{4}{c|}{caption retrieval} & \multicolumn{4}{c}{image retrieval} \\ \hline
                    & R@1      & R@5     & R@10    & Medr    & R@1     & R@5     & R@10    & Medr   \\ \hline
                    & \multicolumn{8}{c}{\textit{1k Test Images}}                                            \\ \hline
VSE++ \cite{faghri2018vse++}               & 64.6     & 90.0    & 95.7    & 1.0     & 52.0    & 84.3    & 92.0    & 1.0    \\ \hline
GXN \cite{gu2018look}                 & 68.5     & -       & \textbf{97.9}    & 1.0     & 56.6    & -       & 94.5    & 1.0    \\ \hline
SCO \cite{huang2018learning}                & 69.9     & 92.9    & 97.5    & -       & 56.7    & \textbf{87.5}    & \textbf{94.8}    & -      \\ \hline
SCAN(t2i) AVG loss \cite{lee2018stacked}  & 70.9     & \textbf{94.5}    & 97.8    & 1.0     & 56.4    & 87.0    & 93.9    & 1.0    \\ \hline
PVSE \cite{song2019polysemous} & 69.2 & 91.6 & 96.6 & - & 55.2 & 86.5 & 93.7 & - \\ \hline
Ours (SGM)           & \textbf{73.4}     & 93.8    & 97.8    & \textbf{1.0}     & \textbf{57.5}    & 87.3    & 94.3    & \textbf{1.0}    \\ \hline
                    & \multicolumn{8}{c}{\textit{5k Test Images}}                                            \\ \hline
VSE++ \cite{faghri2018vse++}              & 41.3     & 71.3    & 81.2    & 2.0     & 30.3    & 59.4    & 72.4    & 4.0    \\ \hline
GXN \cite{gu2018look}                & 42.0     & -       & 84.7    & 2.0     & 31.7    & -       & 74.6    & 3.0    \\ \hline
SCO \cite{huang2018learning}                & 42.8     & 72.3    & 83.0    & -       & 33.1    & 62.9    & 75.5    & -      \\ \hline
*SCAN(t2i) AVG loss \cite{lee2018stacked} & 43.0     & 75.3    & 85.3    & 2.0     & 32.1    & 61.7    & 74.1    & 3.0    \\ \hline
PVSE \cite{song2019polysemous}  & 45.2 & 74.3 & 84.5 & - & 32.4 & 63.0 & 75.0 & - \\ \hline
Ours (SGM)            & \textbf{50.0}     & \textbf{79.3}    & \textbf{87.9}    & \textbf{2.0}     & \textbf{35.3}    & \textbf{64.9}    & \textbf{76.5}    & \textbf{3.0}    \\ \hline
\end{tabular}
\end{center}
\vspace{-0.3cm}
\end{table*}

\begin{figure*}[h]
\vspace{0cm}
\begin{small}

\begin{tabular}{l|p{8cm}|p{8cm}}
\hline
Query & Man with snowboard \textbf{standing next to} another wearing a mask crazy hands & A person \textbf{touching} an elephant \textbf{in front of} a wall. \\ \hline
SGM & \begin{minipage}{0.42\textwidth}\includegraphics[width=80mm, height=12mm]{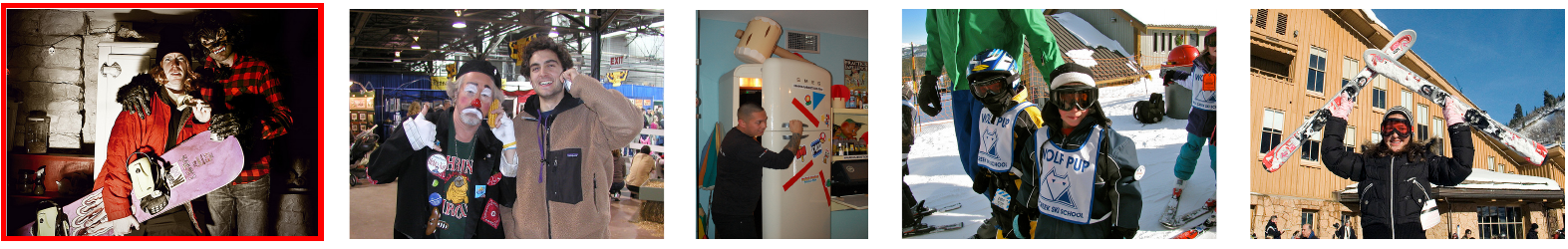} \end{minipage}    &  \begin{minipage}{0.42\textwidth}\includegraphics[width=80mm, height=12mm]{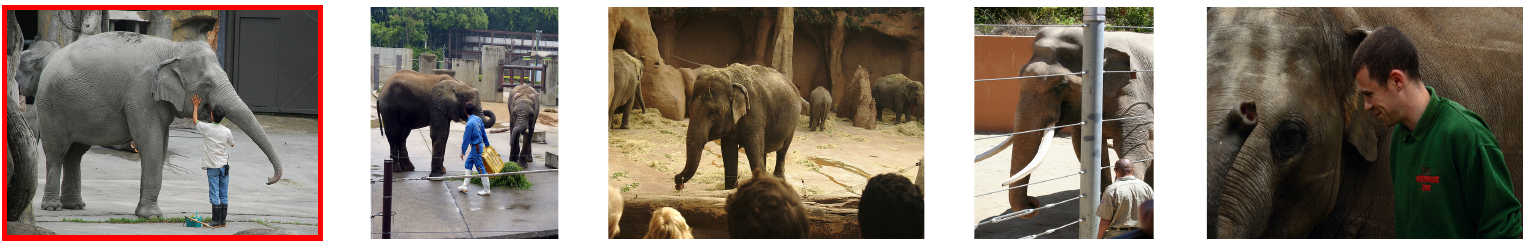} \end{minipage}     \\ \hline
OOM & \begin{minipage}{0.42\textwidth}\includegraphics[width=80mm, height=12mm]{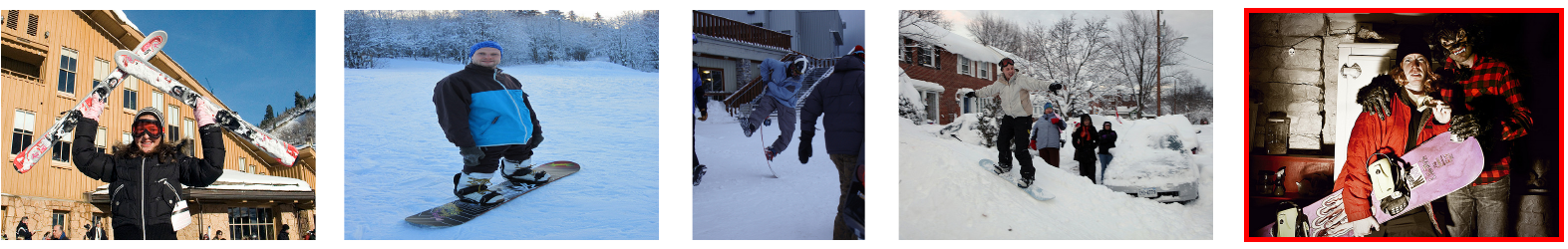} \end{minipage}    &  \begin{minipage}{0.42\textwidth}\includegraphics[width=80mm, height=12mm]{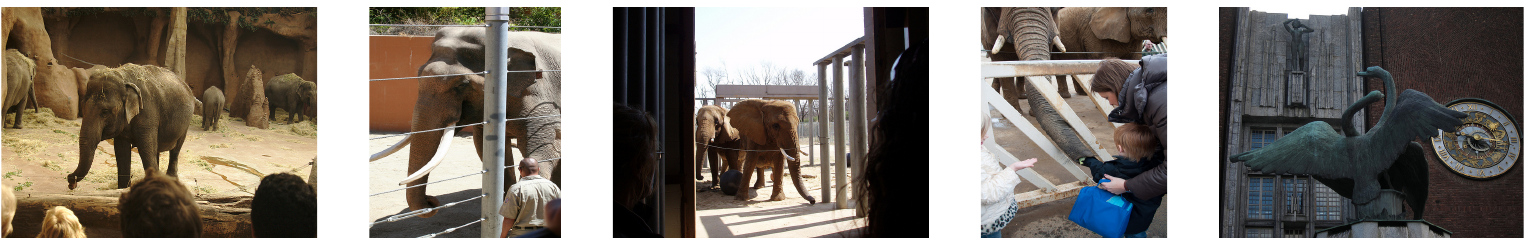} \end{minipage}     \\ \hline

\end{tabular}
\end{small}
\caption{Qualitative top-5 image retrieval results of \textbf{SGM \textbf{vs.} OOM} on MSCOCO. Images with red bounding box are the ground-truth.}
\label{fig:q1}
\end{figure*}

\begin{figure*}[h]
\centering\includegraphics[width=18cm, height=4cm]{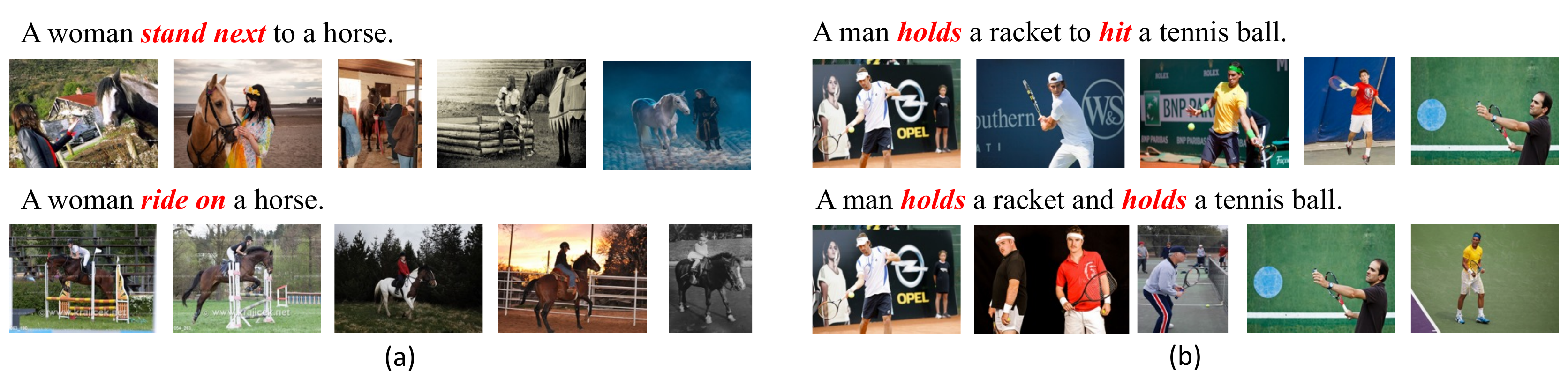}
\caption{Comparison of top-5 retrieved results before and after modifying the relationship words in queries.}
\label{fig:q2}

\end{figure*}

\subsection{Ablation Study}
To justify the importance of relationships for image-text retrieval, we evaluate different variants of our proposed framework in Table \ref{table:ablation_flickr}. \textbf{SGM} is the full model of scene graph matching that contains relationships in both two modalities, and \textbf{OOM} is the model only considering objects matching. \textbf{OOM w VRel} and \textbf{OOM w TRel} stand for adding visual relationships and textual relationships to \textbf{OOM}, respectively. \textbf{OOM w/o TCxt} discards not only the relationships but also textual context, which means words are encoded in isolation rather than word-order bi-GRU. The best results in each column are in bold.

\textbf{Impact of Relationships.} From Table \ref{table:ablation_flickr}, one can find that all other models outperform \textbf{OOM \textbf{w/o} TCxt} that only uses isolated elements for matching. It indicates that associations between objects are essential for image-text retrieval. Comparing \textbf{SGM} with \textbf{OOM}, the importance of relationships for image-text retrieval is revealed. By adding relationship information in both modalities, the performance has enjoyed obvious improvements (especially under the metric R@1) in both tasks of image retrieval and caption retrieval. 

\textbf{Better Representation for Retrieval.}
By adding visual relationships into the model, \textbf{OOM w VRel} outperforms \textbf{OOM} in image retrieval, and the same phenomenon also appears in the comparison between \textbf{SGM} and \textbf{OOM w TRel}. When considering the impact of textual relationships, similar contrasts are observed. Comparing \textbf{OOM w TRel} vs. \textbf{OOM}, and \textbf{SGM} vs. \textbf{OOM w VRel}, it shows that incorporating textual relationships is beneficial to caption retrieval. Such results suggest that better representation in one modality can make the samples in the retrieved database more differentiated and helpful to retrieval task in this modality. While for retrieval task in another modality without relationship features, gains can not be guaranteed. When we add relationship features in both modalities and match at object-level and relationship-level respectively, the performance of cross-modal retrieval obtains a large improvement.

\subsection{Comparison with State-of-the-art Methods}
\label{sec: com_STOA}
In this section, we compare our SGM with state-of-the-art models on Flickr30k and MSCOCO. For a fair comparison, all compared models use ResNet for visual feature extraction. We compared our model with VSE++ \cite{faghri2018vse++}, GXN \cite{gu2018look}, SCO \cite{huang2018learning} and SCAN \cite{lee2018stacked}, which covers both global representation based model and local representation based models. VSE++ embeds full image and sentence into an embedding space and matches them. Its contribution is applying hard negatives mining in training and gaining lots of improvements. 
GXN leverages the image-to-text and text-to-image generative models to learn the local grounded features. SCO concentrates on organizing semantic concepts from images into a correct order before matching with the sentence. SCAN emphasizes attending differentially to important visual objects and words by an attention mechanism. PVSE \cite{song2019polysemous} addresses the issues with ambiguous instances (\eg images containing multiple objects) and partial association by using K embeddings and multiple-instance learning framework. Whereas, we explore the role of relationships for image-text retrieval. 
The results of these methods come from their published papers or are implemented with the published code under the same evaluation protocol. 

As shown in Table \ref{sofa_flickr} and Table \ref{sofa_coco}, our model achieves new state-of-arts on both datasets. We significantly outperform all other methods on Flickr30k and MSCOCO 5k test images by a large margin. On Flickr30k test dataset, our model outperforms the best state-of-the-art model by 16.8\% relatively in caption retrieval, and 16.18\% relatively in image retrieval based on R@1. On MSCOCO 5k test images, we improve caption retrieval by 10.62\% relatively and image retrieval by 6.65\% relatively based on R@1. On MSCOCO 1k test images, while our model delivers slightly lower scores than others under some metrics, it yields clearly superior performance against other competitors under the more crucial metric R@1 for retrieval task. Moreover, all local representation based models surpass the global representation based model (VSE++), which demonstrates the effectiveness of detailed matching, and the achievements of our method verify the necessity of considering relationships in image-text retrieval.   

\vspace{-0.2cm}
\subsection{Qualitative Results}
We show some image retrieval examples using \textbf{SGM} and \textbf{OOM} to reveal the importance of relationships for image-text retrieval of a complex scene. Given the same text query, the top-5 image retrieval results on MSCOCO by \textbf{SGM} and \textbf{OOM} are shown in Fig.\ref{fig:q1}. The top-5 retrieved images by \textbf{SGM} not only contain the right objects but also the right relationships between them. Images only contain the right objects won't be ranked at the top by \textbf{SGM}. However, results by \textbf{OOM} may overlook relationships information in queries and images. (More cases are detailed in our supplementary material.)

To prove that our \textbf{SGM} really captures relationships, we use some text queries to retrieve images from MSCOCO test dataset, and then modify a relationship word in the query to retrieve again. Two retrieval results are compared in Fig.\ref{fig:q2}. We can see that after modifying the relationship words in the text query, the relationships in retrieval results have changed a lot, but objects have not changed. It demonstrates our model has indeed captured the relationships so that we perform well in cross-modal retrieval task with a complex scenario. (More cases are detailed in our supplementary material.)

\vspace{-0cm}
\section{Conclusion}
In this work, we proposed a graph matching based model for image-text retrieval in a complex scenario that contains various objects. We discover that not only the objects but also their relationships are important for local detailed image-text matching. To capture both objects and relationships in the images and text, we have represented image and text into the visual scene graph and the textual scene graph, respectively. Then we design the Scene Graph Matching (SGM) model to extract the object-level features and relationship-level features from the graphs by two graph encoders for image-text matching. Due to explicitly modeling relationship information, our method outperforms state-of-the-art methods in image-text retrieval experiments on both Flickr30k and MSCOCO. What's more, qualitative experiments show that our approach can truly capture the relationships and is helpful in the image-text retrieval task. 


\textbf{Acknowledgements.} This work was partially supported by 973 Program under
contract No. 2015CB351802, Natural Science Foundation
of China under contracts Nos.61390511, 61772500, Frontier
Science Key Research Project CAS No.QYZDJ-SSW-JSC009
and Youth Innovation Promotion Association CAS No.2015085.

{\small
\bibliographystyle{ieee}
\bibliography{egbib}

\begin{thebibliography}{10}\itemsep=-1pt

\bibitem{Anderson2016SPICESP}
P.~Anderson, B.~Fernando, M.~Johnson, and S.~Gould.
\newblock Spice: Semantic propositional image caption evaluation.
\newblock In {\em European Conference on Computer Vision}, 2016.

\bibitem{Anderson2018BottomUpAT}
P.~Anderson, X.~He, C.~Buehler, D.~Teney, M.~Johnson, S.~Gould, and L.~Zhang.
\newblock Bottom-up and top-down attention for image captioning and visual
  question answering.
\newblock {\em Proceedings of the IEEE Conference on Computer Vision and
  Pattern Recognition}, pages 6077--6086, 2018.

\bibitem{faghri2018vse++}
F.~Faghri, D.~J. Fleet, J.~R. Kiros, and S.~Fidler.
\newblock Vse++: Improving visual-semantic embeddings with hard negatives.
\newblock In {\em Proceedings of the British Machine Vision Conference}, 2018.

\bibitem{frome2013devise}
A.~Frome, G.~S. Corrado, J.~Shlens, S.~Bengio, J.~Dean, T.~Mikolov, et~al.
\newblock Devise: A deep visual-semantic embedding model.
\newblock In {\em Advances in Neural Information Processing Systems}, pages
  2121--2129, 2013.

\bibitem{gu2018look}
J.~Gu, J.~Cai, S.~R. Joty, L.~Niu, and G.~Wang.
\newblock Look, imagine and match: Improving textual-visual cross-modal
  retrieval with generative models.
\newblock In {\em Proceedings of the IEEE Conference on Computer Vision and
  Pattern Recognition}, pages 7181--7189, 2018.

\bibitem{he2016deep}
K.~He, X.~Zhang, S.~Ren, and J.~Sun.
\newblock Deep residual learning for image recognition.
\newblock In {\em Proceedings of the IEEE Conference on Computer Vision and
  Pattern Recognition}, pages 770--778, 2016.

\bibitem{huang2017instance}
Y.~Huang, W.~Wang, and L.~Wang.
\newblock Instance-aware image and sentence matching with selective multimodal
  lstm.
\newblock In {\em Proceedings of the IEEE Conference on Computer Vision and
  Pattern Recognition}, pages 2310--2318, 2017.

\bibitem{huang2018learning}
Y.~Huang, Q.~Wu, C.~Song, and L.~Wang.
\newblock Learning semantic concepts and order for image and sentence matching.
\newblock In {\em Proceedings of the IEEE Conference on Computer Vision and
  Pattern Recognition}, pages 6163--6171, 2018.

\bibitem{Johnson2018ImageGF}
J.~Johnson, A.~Gupta, and L.~Fei-Fei.
\newblock Image generation from scene graphs.
\newblock {\em Proceedings of the IEEE Conference on Computer Vision and
  Pattern Recognition}, pages 1219--1228, 2018.

\bibitem{johnson2015image}
J.~Johnson, R.~Krishna, M.~Stark, L.-J. Li, D.~Shamma, M.~Bernstein, and
  L.~Fei-Fei.
\newblock Image retrieval using scene graphs.
\newblock In {\em Proceedings of the IEEE Conference on Computer Vision and
  Pattern Recognition}, pages 3668--3678, 2015.

\bibitem{karpathy2015deep}
A.~Karpathy and L.~Fei-Fei.
\newblock Deep visual-semantic alignments for generating image descriptions.
\newblock In {\em Proceedings of the IEEE Conference on Computer Vision and
  Pattern Recognition}, pages 3128--3137, 2015.

\bibitem{karpathy2014deep}
A.~Karpathy, A.~Joulin, and L.~Fei-Fei.
\newblock Deep fragment embeddings for bidirectional image sentence mapping.
\newblock In {\em Advances in Neural Information Processing Systems}, pages
  1889--1897, 2014.

\bibitem{kingma2014adam}
D.~P. Kingma and J.~Ba.
\newblock Adam: A method for stochastic optimization.
\newblock In {\em International Conference on Learning Representations}, 2015.

\bibitem{kiros2014unifying}
R.~Kiros, R.~Salakhutdinov, and R.~S. Zemel.
\newblock Unifying visual-semantic embeddings with multimodal neural language
  models.
\newblock {\em Advances in Neural Information Processing Systems}, 2014.

\bibitem{Klein2015AssociatingNW}
B.~Klein, G.~Lev, G.~Sadeh, and L.~Wolf.
\newblock Associating neural word embeddings with deep image representations
  using fisher vectors.
\newblock {\em Proceedings of the IEEE Conference on Computer Vision and
  Pattern Recognition}, pages 4437--4446, 2015.

\bibitem{krishnavisualgenome}
R.~Krishna, Y.~Zhu, O.~Groth, J.~Johnson, K.~Hata, J.~Kravitz, S.~Chen,
  Y.~Kalantidis, L.-J. Li, D.~A. Shamma, et~al.
\newblock Visual genome: Connecting language and vision using crowdsourced
  dense image annotations.
\newblock {\em International Journal of Computer Vision}, 123(1):32--73, 2017.

\bibitem{lee2018stacked}
K.-H. Lee, X.~Chen, G.~Hua, H.~Hu, and X.~He.
\newblock Stacked cross attention for image-text matching.
\newblock {\em European Conference on Computer Vision}, 2018.

\bibitem{li2017scene}
Y.~Li, W.~Ouyang, B.~Zhou, K.~Wang, and X.~Wang.
\newblock Scene graph generation from objects, phrases and region captions.
\newblock In {\em International Conference in Computer Vision}, pages
  1261--1270, 2017.

\bibitem{Lin2014MicrosoftCC}
T.-Y. Lin, M.~Maire, S.~J. Belongie, L.~D. Bourdev, R.~B. Girshick, J.~Hays,
  P.~Perona, D.~Ramanan, P.~Doll{\'a}r, and C.~L. Zitnick.
\newblock Microsoft coco: Common objects in context.
\newblock In {\em European Conference on Computer Vision}, 2014.

\bibitem{mikolov2013efficient}
T.~Mikolov, K.~Chen, G.~Corrado, and J.~Dean.
\newblock Efficient estimation of word representations in vector space.
\newblock In {\em ICLR}, 2013.

\bibitem{ren2015faster}
S.~Ren, K.~He, R.~Girshick, and J.~Sun.
\newblock Faster r-cnn: Towards real-time object detection with region proposal
  networks.
\newblock In {\em Advances in Neural Information Processing Systems}, pages
  91--99, 2015.

\bibitem{schuster2015generating}
S.~Schuster, R.~Krishna, A.~Chang, L.~Fei-Fei, and C.~D. Manning.
\newblock Generating semantically precise scene graphs from textual
  descriptions for improved image retrieval.
\newblock In {\em Proceedings of the fourth workshop on vision and language},
  pages 70--80, 2015.

\bibitem{Simonyan15}
K.~Simonyan and A.~Zisserman.
\newblock Very deep convolutional networks for large-scale image recognition.
\newblock In {\em International Conference on Learning Representations}, 2015.

\bibitem{socher2014grounded}
R.~Socher, A.~Karpathy, Q.~V. Le, C.~D. Manning, and A.~Y. Ng.
\newblock Grounded compositional semantics for finding and describing images
  with sentences.
\newblock {\em Transactions of the Association for Computational Linguistics},
  2:207--218, 2014.

\bibitem{song2019polysemous}
Y.~Song and M.~Soleymani.
\newblock Polysemous visual-semantic embedding for cross-modal retrieval.
\newblock In {\em Proceedings of the IEEE Conference on Computer Vision and
  Pattern Recognition}, June 2019.

\bibitem{Teney2017GraphStructuredRF}
D.~Teney, L.~Liu, and A.~van~den Hengel.
\newblock Graph-structured representations for visual question answering.
\newblock {\em Proceedings of the IEEE Conference on Computer Vision and
  Pattern Recognition}, pages 3233--3241, 2017.

\bibitem{wang2017adversarial}
B.~Wang, Y.~Yang, X.~Xu, A.~Hanjalic, and H.~T. Shen.
\newblock Adversarial cross-modal retrieval.
\newblock In {\em Proceedings of the ACM international conference on
  Multimedia}, pages 154--162. ACM, 2017.

\bibitem{wang2016learning}
L.~Wang, Y.~Li, and S.~Lazebnik.
\newblock Learning deep structure-preserving image-text embeddings.
\newblock In {\em Proceedings of the IEEE conference on computer vision and
  pattern recognition}, pages 5005--5013, 2016.

\bibitem{Wang2018NeighbourhoodWR}
P.~Wang, Q.~Wu, J.~Cao, C.~Shen, L.~Gao, and A.~v.~d. Hengel.
\newblock Neighbourhood watch: Referring expression comprehension via
  language-guided graph attention networks.
\newblock In {\em Proceedings of the IEEE Conference on Computer Vision and
  Pattern Recognition (CVPR)}, June 2019.

\bibitem{wang2018scene}
Y.-S. Wang, C.~Liu, X.~Zeng, and A.~Yuille.
\newblock Scene graph parsing as dependency parsing.
\newblock {\em Conference of the North American Chapter of the Association for
  Computational Linguistics}, 2018.

\bibitem{wehrmann2018bidirectional}
J.~Wehrmann and R.~C. Barros.
\newblock Bidirectional retrieval made simple.
\newblock In {\em Proceedings of the IEEE Conference on Computer Vision and
  Pattern Recognition}, pages 7718--7726, 2018.

\bibitem{Wu2019ACS}
Z.~Wu, S.~Pan, F.~Chen, G.~Long, C.~Zhang, and P.~S. Yu.
\newblock A comprehensive survey on graph neural networks.
\newblock {\em arXiv preprint arXiv:1901.00596}, 2019.

\bibitem{Xu2017SceneGG}
D.~Xu, Y.~Zhu, C.~B. Choy, and L.~Fei-Fei.
\newblock Scene graph generation by iterative message passing.
\newblock {\em Proceedings of the IEEE Conference on Computer Vision and
  Pattern Recognition}, pages 3097--3106, 2017.

\bibitem{Yang2018AutoEncodingSG}
X.~Yang, K.~Tang, H.~Zhang, and J.~Cai.
\newblock Auto-encoding scene graphs for image captioning.
\newblock In {\em The IEEE Conference on Computer Vision and Pattern
  Recognition (CVPR)}, June 2019.

\bibitem{Yao2018ExploringVR}
T.~Yao, Y.~Pan, Y.~Li, and T.~Mei.
\newblock Exploring visual relationship for image captioning.
\newblock In {\em European Conference on Computer Vision}, 2018.

\bibitem{Young2014FromID}
P.~Young, A.~Lai, M.~Hodosh, and J.~Hockenmaier.
\newblock From image descriptions to visual denotations: New similarity metrics
  for semantic inference over event descriptions.
\newblock {\em Transactions of the Association for Computational Linguistics},
  2:67--78, 2014.

\bibitem{zellers2018neural}
R.~Zellers, M.~Yatskar, S.~Thomson, and Y.~Choi.
\newblock Neural motifs: Scene graph parsing with global context.
\newblock In {\em Proceedings of the IEEE Conference on Computer Vision and
  Pattern Recognition}, pages 5831--5840, 2018.

\bibitem{zhang2018deep}
Y.~Zhang and H.~Lu.
\newblock Deep cross-modal projection learning for image-text matching.
\newblock In {\em European Conference on Computer Vision}, pages 686--701,
  2018.

\end{thebibliography}
}

\end{document}


\title{Supplementary Material: Cross-modal Scene Graph Matching for Relationship-aware Image-Text Retrieval}
\author{Sijin Wang\textsuperscript{1,2}, Ruiping Wang\textsuperscript{1,2}, Ziwei Yao\textsuperscript{1,2}, Shiguang Shan\textsuperscript{1,2}, Xilin Chen\textsuperscript{1,2} \\
$^1$Key Laboratory of Intelligent Information Processing of Chinese Academy of Sciences (CAS), \\Institute of Computing Technology, CAS, Beijing, 100190, China\\
$^2$University of Chinese Academy of Sciences, Beijing, 100049, China \\
{\tt\small \{sijin.wang, ziwei.yao\}@vipl.ict.ac.cn,} {\tt\small \{wangruiping, sgshan, xlchen\}@ict.ac.cn}
}

\maketitle
\ifwacvfinal\thispagestyle{empty}\fi

\begin{figure*}[hb]
\vspace{-0.5cm}
\begin{center}
\begin{tabular}{l|p{8cm}|p{8cm}}
\hline
Query & Person with bananas \textbf{on} head and banana necklace.& A beautiful vase full of flowers and pictures \textbf{next to} it. \\ \hline
SGM & \begin{minipage}{0.42\textwidth}\includegraphics[width=80mm, height=15mm]{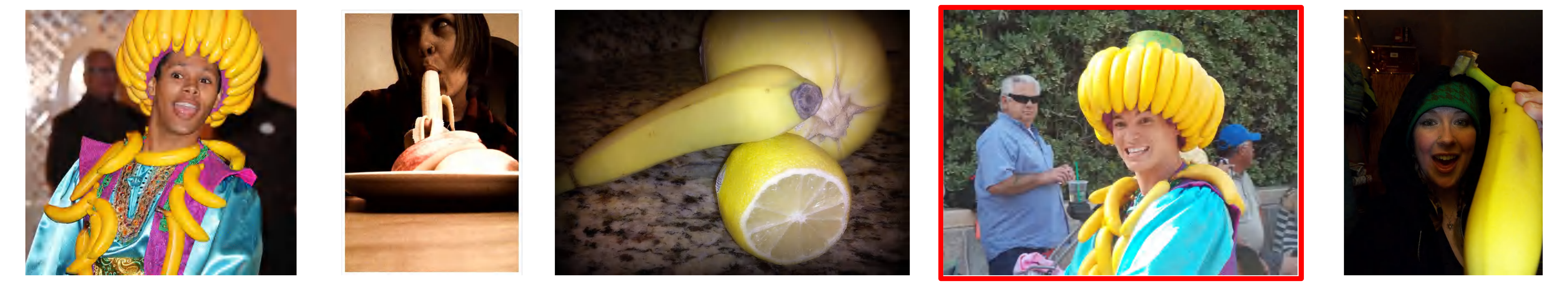} \end{minipage}    &  \begin{minipage}{0.42\textwidth}\includegraphics[width=80mm, height=15mm]{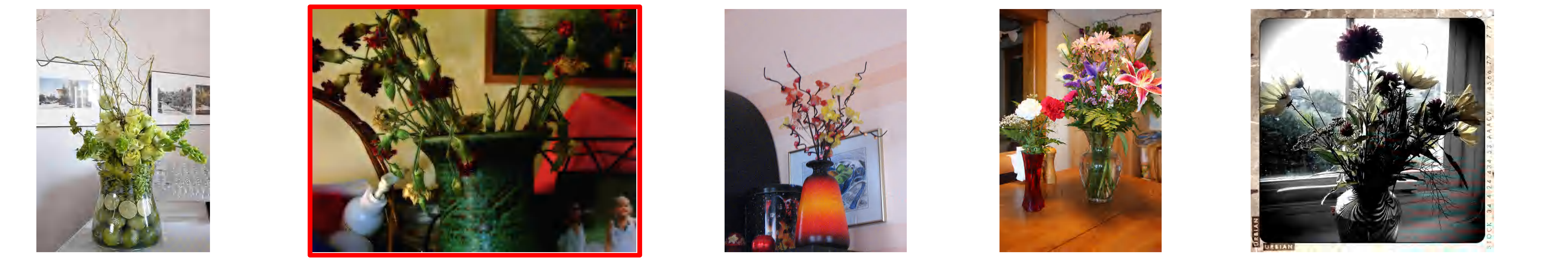} \end{minipage}     \\ \hline
OOM & \begin{minipage}{0.42\textwidth}\includegraphics[width=80mm, height=15mm]{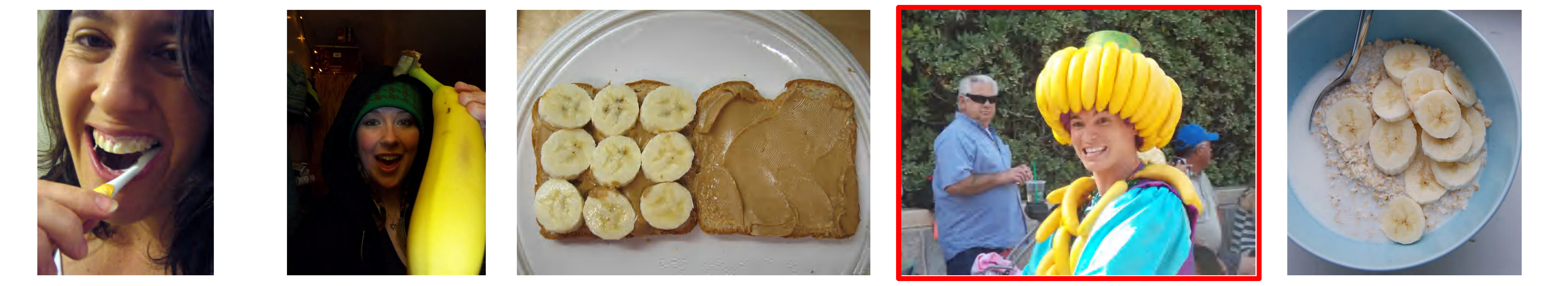} \end{minipage}    &  \begin{minipage}{0.42\textwidth}\includegraphics[width=80mm, height=15mm]{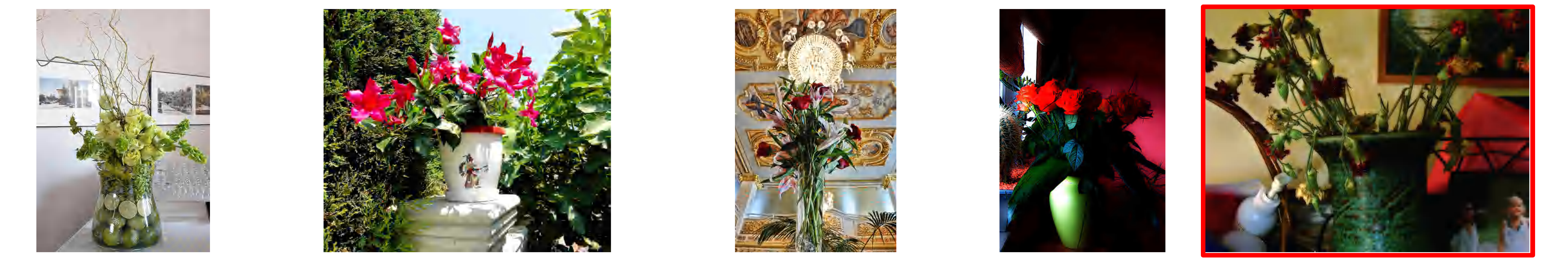} \end{minipage}     \\ \hline

\end{tabular}
\end{center}
\caption{Qualitative image retrieval results of \textbf{SGM \textbf{vs.} OOM} on MSCOCO. Images with red bounding box are the ground-truth.}
\label{fig:q1}
\end{figure*}

\begin{figure*}[hb]
\vspace{+0.5cm}
\begin{center}
\begin{tabular}{l|p{11cm}}
\hline
Query & A young man \textbf{holding} a snow board and a pair of shoes. \\ \hline
SGM & \begin{minipage}{0.42\textwidth}\includegraphics[width=110mm]{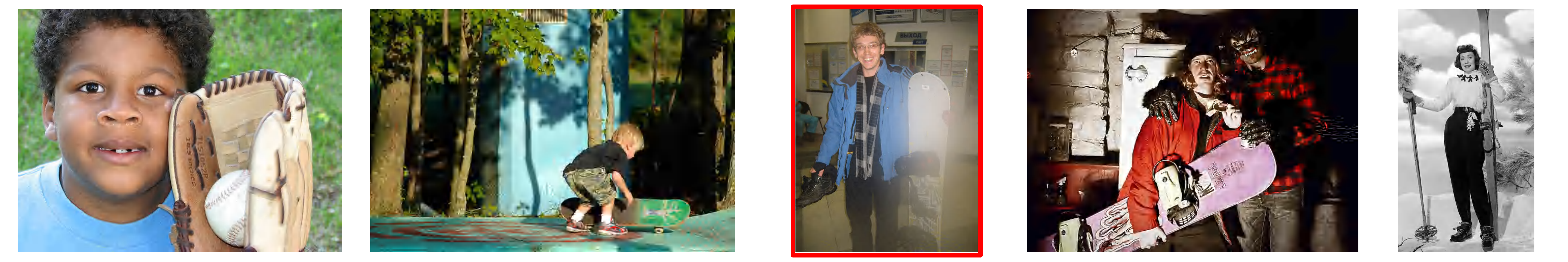} \end{minipage}        \\ \hline
OOM & \begin{minipage}{0.42\textwidth}\includegraphics[width=110mm]{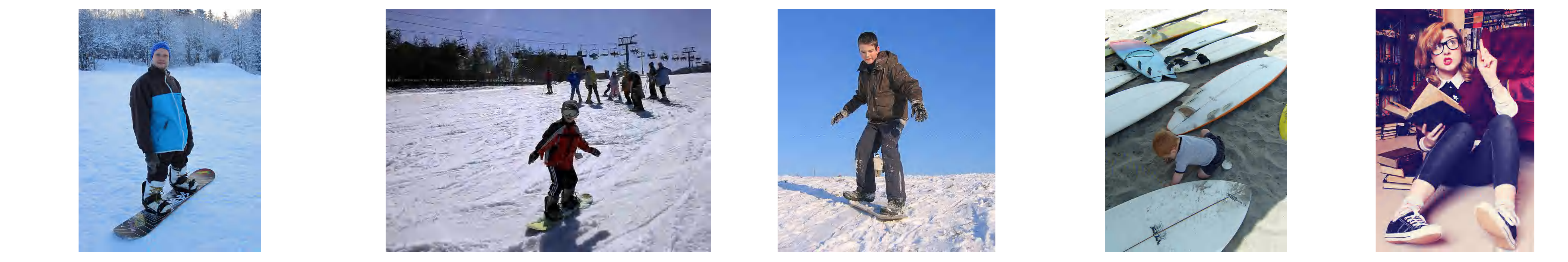} \end{minipage}         \\ \hline

\end{tabular}
\end{center}
\caption{A failure case of SGM and OOM. Image with red bounding box is the ground-truth.}
\label{fig:q2}
\vspace{-0.4cm}
\end{figure*}

We illustrate more qualitative image retrieval results of SGM vs. OOM on MSCOCO in Fig. \ref{fig:q1} and it proves that relationship-aware matching method is better than the method that only addresses the object-level matching. We show a failure case in Fig. \ref{fig:q2}. The failure case shows that SGM sometimes focuses too much on relationships, so how to balance the emphasis on objects and relationships will be in our future work. The OOM model also fails in this case, which only retrieves the images with correct objects but wrong relationships.

\begin{figure*}[hb]
\centering\includegraphics[width=18cm]{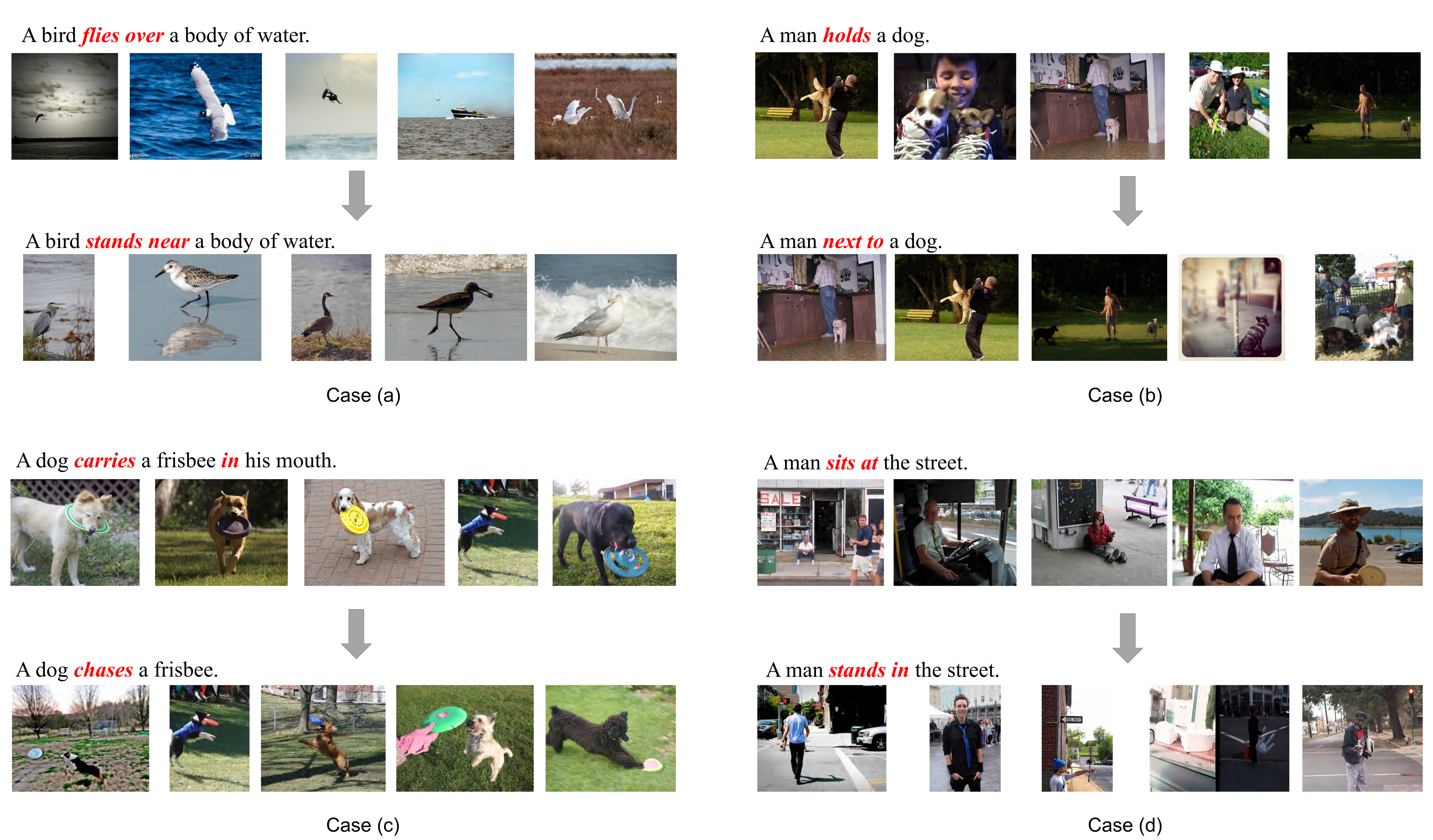}
\vspace{-0.1cm}
\caption{Comparison of top-5 retrieved results before and after modifying the relationship words in queries.}
\label{fig:f3}

\end{figure*}

Then in Fig.\ref{fig:f3}, we show more cases that the SGM has indeed captured the relationships. So when the relationship word in the query is modified, the retrieved results have also changed a lot accordingly.

